%% file: neurips_2023.tex
\documentclass{article}

\usepackage[nonatbib,preprint]{neurips_2023}

\usepackage[utf8]{inputenc} 
\usepackage[T1]{fontenc}    
\usepackage{hyperref}       
\usepackage{url}            
\usepackage{booktabs}       
\usepackage{amsfonts}       
\usepackage{nicefrac}       
\usepackage{microtype}      
\usepackage{xcolor}         

\usepackage{multirow}
\usepackage{longtable}
\usepackage{adjustbox}
\usepackage{xspace}  
\usepackage{graphicx}
\newcommand{\xhdr}[1]{\vspace{1.7mm}\noindent{{\bf #1.}}}
\usepackage[linesnumbered,ruled,vlined]{algorithm2e}
\newcommand{\hrulealg}[0]{\vspace{1mm} \hrule \vspace{1mm}}
\let\oldnl\nl
\newcommand{\nlnonumber}{\renewcommand{\nl}{\let\nl\oldnl}}
\definecolor{darkpastelgreen}{rgb}{0.01, 0.75, 0.24}
\usepackage{bbding}
\usepackage{pifont}
\newcommand{\greencheck}{{\color{darkpastelgreen}\CheckmarkBold}}
\newcommand{\redmark}{{\color{red}\ding{55}}}
\input{math_commands}

\newcommand{\methodname}{\textsc{Plato}\xspace}
\newcommand{\numbaselines}{13\xspace}
\newcommand{\numdggndatasets}{6\xspace}
\newcommand{\numdsimndatasets}{4\xspace}
\newcommand{\smallestfeaturenum}{12,932\xspace}

\newcommand{\std}[1]{\scriptsize{$\pm$#1}}
\newcommand{\numconfigurations}{500\xspace}
\newcommand{\bestimprovement}{10.19\%\xspace}

\newcommand{\numnodes}{108,447 }

\newcommand{\numedges}{3,066,156 }

\newcommand{\numrelationtypes}{99 }

\title{Enabling tabular deep learning when $d \gg n$ with an auxiliary knowledge graph}

\author{
Camilo Ruiz$^{1,2,\ast}$, Hongyu Ren$^{1,\ast}$, Kexin Huang$^{1}$, Jure Leskovec$^{1}$ \\
$^{1}$Department of Computer Science, Stanford University \\
$^{2}$Department of Bioengineering, Stanford University \\
$^{*}$Equal contribution \\
\texttt{\{caruiz, hyren, kexinh, jure\}@cs.stanford.edu} \\
}

\begin{document}
\maketitle
\input{000_abstract}
\input{100_intro}
\input{500_related}
\input{300_method}
\input{400_experiment}
\input{600_conclusion}

\bibliographystyle{plain}
\bibliography{refs}

\input{700_appendix}
\end{document}

%% file: math_commands.tex
\usepackage{amsmath,amsfonts,bm}

\def\Figref#1{Figure~\ref{#1}}

\def\eqref#1{equation~\ref{#1}}

\def\1{\bm{1}}

\DeclareMathAlphabet{\mathsfit}{\encodingdefault}{\sfdefault}{m}{sl}
\SetMathAlphabet{\mathsfit}{bold}{\encodingdefault}{\sfdefault}{bx}{n}

\newcommand{\softmax}{\mathrm{softmax}}



%% file: 000_abstract.tex
\begin{abstract}
Machine learning models exhibit strong performance on datasets with abundant labeled samples.
However, for tabular datasets with extremely high $d$-dimensional features but limited $n$ samples (i.e. $d \gg n$), machine learning models struggle to achieve strong performance due to the risk of overfitting.
Here, our key insight is that there is often abundant, auxiliary domain information describing input features which can be structured as a heterogeneous knowledge graph (KG).
We propose \methodname, a method that achieves strong performance on tabular data with $d \gg n$ by using an auxiliary KG describing input features to regularize a multilayer perceptron (MLP).
In \methodname, each input feature corresponds to a node in the auxiliary KG.
In the MLP’s first layer, each input feature also corresponds to a weight vector.
\methodname is based on the inductive bias that two input features corresponding to similar nodes in the auxiliary KG should have similar weight vectors in the MLP's first layer.
\methodname captures this inductive bias by inferring the weight vector for each input feature from its corresponding node in the KG via a trainable message-passing function.
Across 6 $d \gg n$ datasets, \methodname outperforms 13 state-of-the-art baselines by up to \bestimprovement.

\end{abstract}

%% file: 100_intro.tex
\begin{figure}[t]
    \centering
    \includegraphics[width = .98\textwidth]{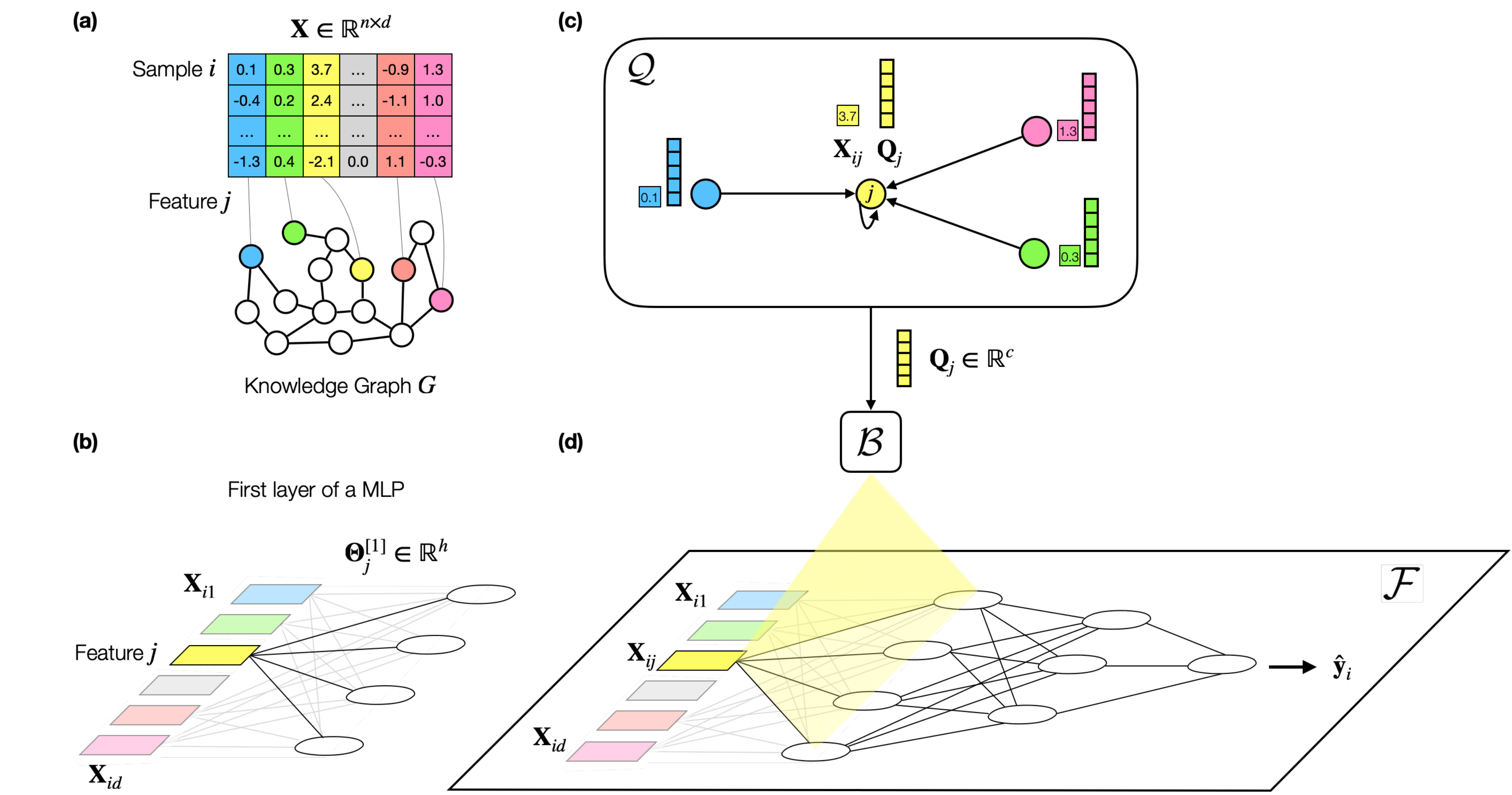}
    \caption{\textbf{\methodname is a method that uses auxiliary domain information describing input features to regularize a multilayer perceptron (MLP) and achieve strong performance on tabular data with $d \gg n$.}
    \textbf{(a)} In \methodname, each input feature $j$ corresponds to a node in an auxiliary KG of domain information.
    \textbf{(b)} In the first layer of a MLP with $h$ hidden units, each input feature $j$ corresponds to a vector of weights $\mathbf{\Theta}^{[1]}_j \in \mathbb{R}^h$ such that the weight vectors of all $d$ features compose the weight matrix $\mathbf{\Theta}^{[1]} \in \mathbb{R}^{d \times h}$. \methodname is based on the inductive bias that, if two input features $j$ and $k$ correspond to similar nodes in the auxiliary KG, they should have similar weight vectors $\mathbf{\Theta}^{[1]}_j$ and $\mathbf{\Theta}^{[1]}_k$ in the MLP.
    \textbf{(c,d)} \methodname captures this inductive bias by inferring the weight vector for each input feature $j$ from its corresponding node in the KG. A trainable message-passing function $\mathcal{Q}$ creates a low-dimensional embedding $\mathbf{Q}_j \in \mathbb{R}^{c}$ for each input feature $j$.
    A neural network $\mathcal{B}$ that is shared across all input features then infers the weight vector $\mathbf{\Theta}^{[1]}_j$ corresponding to input feature $j$ from $\mathbf{Q}_j$. Input features with similar embeddings produce similar weight vectors, regularizing the MLP.
    }
    \label{fig:figure1}
    \vspace{-6mm}
\end{figure}

\section{Introduction}
\label{sec:intro}

Machine learning models have reached state-of-the-art performance in domains with abundant labeled data like computer vision~\cite{wortsman2022model,deng2009imagenet} and natural language processing~\cite{wang2018glue,devlin2018bert,ramesh2022hierarchical}. However, for tabular datasets in which the number $ d $ of features vastly exceeds the number $ n $ of samples, machine learning models struggle to achieve strong performance~\cite{hastie2009elements,liu2017deep}. Crucially, many tabular datasets from scientific domains~\cite{guyon2004result, iorio2016landscape, yang2012genomics, garnett2012systematic,gao2015high, kasieczka2021lhc} have high-dimensional features but limited labeled samples due to the high time and labor costs of experiments. For these and other tabular datasets with $ d \gg n $, the performance of machine learning models is currently limited. 

The key challenge for machine learning models when $d \gg n$ is the risk of overfitting. Indeed, deep models can have a large number of trainable weights, yet training is limited by the comparatively small number of labeled samples.  As a result, tabular deep learning approaches so far have focused on data-rich regimes with far more samples than features ($ n \gg d $)~\cite{grinsztajn2022tree, gorishniy2021revisiting, shwartz2022tabular}. In the low-data regime with far more features than samples ($ d \gg n $), the dominant approaches are still statistical methods~\cite{hastie2009elements}. These statistical methods reduce the dimensionality of the input space~\cite{abdi2010principal,liu2017deep,van2008visualizing,van2009dimensionality}, select features~\cite{tibshirani1996regression,climente2019block,freidling2021post,meier2008group}, impose regularization penalties on parameter magnitudes~\cite{marquardt1975ridge}, or use ensembles of weak tree-based models~\cite{friedman2001greedy, chen2016xgboost, ke2017lightgbm, lou2017bdt, prokhorenkova2018catboost}.

Here, we present a novel problem setting and framework that enables tabular deep learning when $ d \gg n $ (Figure \ref{fig:figure1}). Our key insight is that there is often abundant, auxiliary domain information describing input features which can be structured as a heterogeneous knowledge graph (KG). We propose a novel problem setting in which each input feature of a tabular dataset corresponds to a node in an auxiliary KG (Figure \ref{fig:figure1}a). To represent diverse domain information describing the input features, the KG contains feature and non-feature nodes as well as multiple node and edge types. For example, consider a tabular medical dataset in which each row is a cancer patient, each column is a gene, and each value is the amount of a gene in the patient’s tumor. For this tabular dataset, there exists an auxiliary KG with each gene (\emph{i.e.} input feature) as a node. Each gene node has edges to other gene nodes (\emph{i.e.} other feature nodes) with diverse edge types like ``activates'' or ``inhibits.'' Each gene node also has edges to other nodes (\emph{i.e.} non-feature nodes) representing the gene's function in the body like ``heart rate''. Finally, the function nodes (\emph{i.e.} non-feature nodes) have edges to each other representing their anatomical relationships like ``heart rate''-``part of''-''cardiac system''. Note that the KG does \emph{not} capture the relationships between \emph{input data samples} but instead captures the relationships between \emph{input features} and other domain information.

Within our novel problem setting, we propose \methodname, a method that enables deep learning for tabular data with $d \gg n$ by using an auxiliary KG describing input features (\Figref{fig:figure1}). \methodname achieves strong performance by using the auxiliary KG to regularize a multilayer perceptron (MLP). In \methodname, each input feature corresponds to a node in the auxiliary KG (\Figref{fig:figure1}a). In the first layer of the MLP, each input feature also corresponds to a weight vector such that the weight vectors of all features collectively compose the weight matrix (\Figref{fig:figure1}b). \methodname is based on the inductive bias that two input features which correspond to similar nodes in the KG should have similar weight vectors in the first layer of the MLP. \methodname captures this inductive bias by inferring the weight vector for a feature from its corresponding node in the auxiliary KG with a trainable message-passing function (\Figref{fig:figure1}c,d). Inferring the weights in the MLP's first layer also leads to a drastic reduction in the number of trainable weights, since most weights in a MLP are usually in the first layer when $d \gg n$.

We exhibit $\methodname$’s performance on \numdggndatasets $d \gg n$ tabular datasets with \numbaselines state-of-the-art baselines spanning dimensionality reduction, feature selection, statistical models, graph regularization, parameter-inference, and tabular deep learning. Following a rigorous evaluation protocol from the tabular deep learning literature~\cite{grinsztajn2022tree, gorishniy2021revisiting}, \methodname outperforms the prior state-of-the-art on all \numdggndatasets datasets by up to \bestimprovement. Ablation studies demonstrate the importance of \methodname's trainable message-passing, the importance of non-feature nodes in the KG, and \methodname's robustness to missing edges in the KG. Ultimately, \methodname enables deep learning for tabular data with $d \gg n$ by using an auxiliary KG describing the input features.

%% file: 500_related.tex
\section{Related Work}
\xhdr{Tabular deep learning methods}
In contrast to \methodname's setting, tabular deep learning methods have been developed for settings with far more samples than features (\emph{i.e.} $n \gg d$). Recent tabular deep learning benchmarks ignore datasets with high numbers of features and low numbers of samples~\cite{grinsztajn2022tree, gorishniy2021revisiting, shwartz2022tabular}. In the $n \gg d$ setting, various categories of deep tabular models have been benchmarked. We compare \methodname to several of the state-of-the-art models. First, decision tree models like NODE~\cite{popov2019neural} make decision trees differentiable to enable gradient-based optimization~\cite{hazimeh2020tree, kontschieder2015deep, yang2018deep}. Second, tabular transformer architectures use an attention mechanism to select and learn interactions among features. These include TabNet~\cite{arik2021tabnet}, TabTransformer~\cite{huang2020tabtransformer}, FT-Transformer~\cite{gorishniy2021revisiting}, and others~\cite{song2019autoint, somepalli2021saint, kossen2021self}. 

\xhdr{$d \gg n$ methods}
For \methodname's setting in which $d \gg n$, various tabular machine learning approaches exist~\cite{hastie2009elements}. First, dimensionality reduction techniques like PCA~\cite{abdi2010principal} aim to reduce the dimensionality of the input data while preserving as much of the the variance in the data as possible~\cite{liu2017deep,van2008visualizing,van2009dimensionality}. Second, feature selection approaches select a parsimonious set of features, leading to a smaller feature space. Feature selection approaches include LASSO~\cite{tibshirani1996regression} and its variants~\cite{climente2019block,freidling2021post,meier2008group}. For feature selection with deep learning, Stochastic Gates~\cite{yamada2020feature} are among the best performing of many variants~\cite{balin2019concrete,lu2018deeppink}. Finally, tree-based models like XGBoost learn ensembles of weak decision trees models to make an overall prediction~\cite{friedman2001greedy, chen2016xgboost, ke2017lightgbm, prokhorenkova2018catboost}.

\xhdr{Weight inference} Using one network to infer the weights of another has been studied extensively~\cite{denil2013predicting,schmidhuber1992learning,bengio1990learning}. For example, \cite{ha2016hypernetworks} infers the weights in all layers of a sequential model (\emph{i.e.} RNN, LSTM) by using information about the weights' structure. Diet Networks~\cite{romero2016diet} infer weights by hand-crafting prior information about the input features or using random projections. By contrast, \methodname infers the weights in a MLP from prior information describing the input features in an auxiliary KG. \methodname's weight inference uniquely captures the inductive bias that two input features corresponding to similar nodes in a KG should have similar corresponding weight vectors in the first layer of a MLP (\Figref{fig:figure1}). 

\xhdr{Graph regularization}
Graph regularization approaches regularize the weights of a linear model based on a simple graph between input features. The graph is typically constructed from the tabular data based on covariance relationships. Approaches then add a regularization penalty to the loss function which forces the weights of the linear model to vary smoothly over the corresponding feature nodes in the graph. State-of-the-art methods include GraphNet~\cite{grosenick2013interpretable} and Network-Constrained LASSO~\cite{li2008network} which are based on a Laplacian regularization~\cite{smola2003kernels,ando2006learning} as well as Network LASSO~\cite{hallac2015network} which generalizes the Group LASSO~\cite{yuan2006model} to a network setting. \methodname differs from graph regularization approaches in two key ways. First, \methodname's KG includes both feature and non-feature nodes and multiple edge types, thereby modeling diverse, prior domain information that is missing in graph regularization approaches. Second, \methodname infers the weights of a deep non-linear model (\emph{i.e.} a MLP) rather than adding a regularization penalty to a loss, representing a new regularization mechanism.

\xhdr{Knowledge graph methods}
Existing KG approaches are designed for tasks directly on the graph like link prediction~\cite{wang2017knowledge, trouillon2016complex,wang2014knowledge,yang2014embedding, d2021injecting}. By contrast, \methodname does not make any predictions on the KG. Instead, \methodname makes predictions on a separate, tabular dataset by using the KG as prior information describing the features and domain. Graph classification methods are also not relevant (Appendix \ref{app:graph_classify}).

%% file: 300_method.tex
\section{\methodname}

\methodname is a machine learning method for tabular datasets with $d \gg n$ and an auxiliary knowledge graph (KG) with input features as nodes (Section \ref{sec:problem-setting}). The key insight of \methodname is that there often exists abundant domain information describing input features which can be structured as an auxiliary KG $G$ (\Figref{fig:figure1}a). \methodname uses the auxiliary KG to regularize a multilayer perceptron (MLP) and achieve strong performance on tabular data when $d \gg n$.

\subsection{Problem setting}\label{sec:problem-setting}
Consider a tabular dataset $ \mathbf{X} \in \mathbb{R}^{n \times d}$ with labels $ \mathbf{y} \in \mathbb{R}^{n} $ and far more $d$ features than $n$ samples such that $ d \gg n $. The goal is to train a model $ \mathcal{F} $ to predict labels $ \mathbf{\hat{y}}$ from the input $ \mathbf{X}$. \methodname assumes the existence of an auxiliary knowledge graph $ G = (V, E)$ with $ |V| $ nodes and $ |E| $ edges such that each input feature $j$ corresponds to a node in $ G $. Formally, $\forall j \in \{1, \ldots, d \}\text{, }\exists v \in V \text{ s.t. } j \mapsto v$, as shown in \Figref{fig:figure1}a. $ G $ also contains additional nodes which represent broader knowledge describing the domain. The edges in $G$ are (head node, relation type, tail node) triplets.

\subsection{\methodname's inductive bias}
In \methodname, each input feature $j$ corresponds to a node in the auxiliary KG (\Figref{fig:figure1}a). In the first layer of a MLP with $h$ hidden units, each input feature $j$ also corresponds to a weight vector $\mathbf{\Theta}^{[1]}_j \in \mathbb{R}^h$ such that the weight vectors of all features collectively compose the weight matrix $\mathbf{\Theta}^{[1]} \in \mathbb{R}^{d \times h}$ (\Figref{fig:figure1}b). \methodname is based on the inductive bias that two input features $j$ and $k$ which correspond to similar nodes in the KG should have similar weight vectors $\mathbf{\Theta}^{[1]}_j$ and $\mathbf{\Theta}^{[1]}_k$ in the first layer of the MLP. PLATO captures this inductive bias by inferring the weight vector for a feature from its corresponding node in the auxiliary KG with a trainable message-passing function (\Figref{fig:figure1}c,d).

\subsection{\methodname overview}
\methodname has four key steps. First, \methodname uses a self-supervised objective on the auxiliary KG to pretrain an embedding for each input feature (Section \ref{sec:pretrain}). Second, \methodname updates each feature embedding with a trainable message-passing function that is trained on the supervised loss objective for the tabular data (Section \ref{sec:adapt-feat-embeds}, \Figref{fig:figure1}c). Third, \methodname infers the weights in the first layer of the MLP from the feature embeddings with a small neural network that is shared across input features (Section \ref{sec:shallow-decoder}, \Figref{fig:figure1}d). Finally, the MLP predicts the label for the input sample.

\subsection{Pretraining feature embeddings with self-supervision on the knowledge graph}\label{sec:pretrain}
First, \methodname learns general prior information about each input feature $j$ from the auxiliary KG $G$. \methodname represents the general prior information about each input feature $j$ as a low-dimensional embedding $\mathbf{M}_j \in \mathbb{R}^c$. Since each input feature $j$ corresponds to a node in $G$, \methodname can learn $\mathbf{M}_j$ by learning an embedding for the corresponding feature node in $G$. Any self-supervised node embedding method on $G$ can be used within $\methodname$'s framework.

\xhdr{Formal notation}
Formally, \methodname uses self-supervision on $ G $ to pretrain an embedding for each input feature according to
\begin{equation}\label{eq:pretrain} \mathbf{M} = \mathcal{H}(G). \end{equation}
$ \mathbf{M} \in \mathbb{R}^{d \times c}$ is the matrix of all feature embeddings. $ \mathcal{H} $ is a self-supervised node embedding method. We refer to Eq. (\ref{eq:pretrain}) as pretraining since only the auxiliary KG $G$ is used but the tabular data $\mathbf{X}$, $\mathbf{y}$ is ignored. After pretraining, the feature embeddings $ \mathbf{M} $ are fixed.

For $\mathcal{H}$, we choose ComplEx as it is a prominent and highly scalable KG node embedding method~\cite{trouillon2016complex}. ComplEx uses a self-supervised objective which learns an embedding for each node in $G$ by classifying whether a proposed edge exists in $G$. ComplEx's proposed edges include both feature nodes and other nodes in $G$, thereby integrating prior information about the input features and the broader domain. We also test KG embedding methods DistMult~\cite{yang2014embedding} and TransE~\cite{wang2014knowledge} in Appendix \ref{sec:alternative-kg-embed}.

\subsection{Updating feature embeddings with a message-passing function trained on tabular data}\label{sec:adapt-feat-embeds}

\methodname next updates each feature embedding with a trainable message-passing function that is trained on the supervised loss for the tabular data~(\Figref{fig:figure1}c). During message-passing, \methodname updates the embedding of each input feature to be a weighted aggregation of it's neighbors' embeddings.

\xhdr{Formal notation} Formally, \methodname uses a message-passing function $ \mathcal{Q}$ on the KG to update each pre-trained feature embedding $\mathbf{M}_j \in \mathbb{R}^c$ to feature embedding $\mathbf{Q}_j \in \mathbb{R}^c$ according to 

\begin{subequations}\label{eq:q-message-pass}
\begin{equation} \label{eq_q}\mathbf{Q} = \mathcal{Q}(\mathbf{M}, G, \mathbf{X}_i; \mathbf{\Pi}).\tag{\ref{eq:q-message-pass}}\end{equation}
As input, the message-passing function considers the pre-trained feature embeddings $\mathbf{M}$, the knowledge graph $G$, and the sample value $\mathbf{X}_i$.
$\mathcal{Q}$ uses an attention mechanism which considers the sample value $\mathbf{X}_i$.
The only trainable weights in $\mathcal{Q}$ are in the attention mechanism and are $\mathbf{\Pi}$.

\xhdr{The message passing network $\mathcal{Q}$}
Let $ \mathbf{Q}_j^{[r]} $ be the embedding of input feature $ j $ after round $ r \in \{1, ..., R\}$ of message passing. For each input feature $ j $, $ \mathcal{Q}$ first initializes the updated feature embedding to the pretrained feature embedding. 
\begin{equation} \mathbf{Q}_j^{[0]} = \mathbf{M}_j. \end{equation} 
$ \mathcal{Q} $ then conducts $ R $ rounds of message passing. In each round of message passing, the feature embedding $\mathbf{Q}_j^{[r]}$ is updated from the feature embedding of each neighbor $k$ in the prior round $\mathbf{Q}_k^{[r-1]}$ and its own feature embedding in the prior round $\mathbf{Q}_j^{[r-1]}$. The ``message'' being passed is the embedding of each feature from the prior round.
\begin{equation} \mathbf{Q}_j^{[r]} = \sigma \bigg[ \overbrace{\beta (\sum_{k \in N_j} \alpha_{ijk} \mathbf{Q}_k^{[r-1]} )}^{\text{Weighted messages from neighbors}} + \underbrace{{(1 - \beta) \mathbf{Q}_j^{[r-1]}}}_{\text{Weighted message from self}} \bigg].\end{equation}

$ \mathbf{\sigma}$ is an optional nonlinearity. $ N_j $ are the neighbors of feature node $ j $ in $ G $. 

During message-passing, $\mathcal{Q}$ uses two scalar values $\beta \in \mathbb{R}$ and $\alpha_{ijk} \in \mathbb{R}$ to control the weights of messages. First, $\mathcal{Q}$ uses hyperparameter $\beta \in \mathbb{R}$ to control the weight of the messages aggregated from the feature node's neighbors vs. from the feature node itself. Second, $\mathcal{Q}$ calculates an attention coefficient $\alpha_{ijk} \in \mathbb{R}$ to allow distinct nodes in the same neighborhood to have distinct weights. The coefficient $\alpha_{ijk}$ specifies the weight of the message between feature $j$ and neighbor $k$ for sample $i$.

After $ R $ rounds of message-passing, the updated feature embeddings $ \mathbf{Q}_j$ are set.
\begin{equation} \mathbf{Q}_j = \mathbf{Q}_j^{[R]}.\end{equation}

\xhdr{The attention coefficient}
\methodname's attention coefficient $\alpha_{ijk}$ is inspired by~\cite{velivckovic2017graph} in which node attributes are used to calculate the weight of a message between neighboring nodes. For a sample $i$ in \methodname, the node attributes for features $j$ and $k$ are their sample values $\mathbf{X}_{ij} \in \mathbb{R}$ and $\mathbf{X}_{ik} \in \mathbb{R}$. \methodname thus uses the sample values $\mathbf{X}_{ij}$ and $\mathbf{X}_{ik}$ to calculate the attention coefficient. The attention coefficient $e_{ijk}$ indicates the importance of node $j$ to node $k$ for sample $i$.
\begin{equation} e_{ijk} = \mathcal{A}(\mathbf{X}_{ij}, \mathbf{X}_{ik}; \mathbf{\Pi}).\end{equation}
 $\mathcal{A}$ is a shallow neural network parameterized by $\mathbf{\Pi}$ that is shared across samples and features. The number of trainable weights in $\mathbf{\Pi}$ is small since the input of $\mathcal{A}$ is $\mathbb{R}^2$ and the output of $\mathcal{A}$ is a scalar $\mathbb{R}$.
 
 To make the attention coefficients comparable across different nodes, \methodname normalizes the attention coefficients with a softmax function across the neighbors $N_j$ of node $j$.
\begin{equation} \alpha_{ijk} = \softmax_k(e_{ijk}) = \frac{\exp{(e_{ijk})}}{\sum_{t \in N_j}\exp{(e_{ijt}})} .\end{equation}

\end{subequations}

\begin{algorithm}[t]
\SetAlgoLined
\DontPrintSemicolon
\KwIn{A data sample $ \mathbf{X}_i \in \mathbb{R}^{d}$, a knowledge graph $G$ containing each input feature in $\mathbf{X}$ as a node, a matrix of input feature embeddings $\mathbf{M} \in \mathbb{R}^{d \times c}$ pre-trained over $G$.}
\KwOut{A predicted label $ \mathbf{\hat{y}}_i \in \mathbb{R} $.}

\hrulealg

Use a trainable message-passing function $\mathcal{Q}$ to update the pre-trained feature embeddings:

\nlnonumber
$\mathbf{Q} = \mathcal{Q}(\mathbf{M}, G, \mathbf{X}_i; \mathbf{\Pi})$, $\mathbf{Q}_j \in \mathbb{R}^c$, $\mathbf{Q} \in \mathbb{R}^{d \times c}$

Infer the weight vector in the first layer of a MLP that corresponds to an input feature $j$ with a neural network $\mathcal{B}$:

\nlnonumber
$ \mathbf{\hat{\Theta}}^{[1]}_j  = \mathcal{B}(\mathbf{Q}_j | \mathbf{X}_i ; \mathbf{\Phi}) $, $\mathbf{\hat{\Theta}}^{[1]}_j \in \mathbb{R}^{h}$

Repeat to infer the weight vectors corresponding to all input features by sharing the neural network $\mathcal{B}$:

\nlnonumber
$\mathbf{\hat{\Theta}}^{[1]} \in \mathbb{R}^{d \times h}$

Concatenate the first layer inferred weights with the trainable weights in the rest of the MLP layers:

\nlnonumber
$\mathbf{\hat{\Theta}} = \{\mathbf{\hat{\Theta}}^{[1]} | \mathbf{X}_i\}  \cup \{ \mathbf{\Theta}^{[2]}, \ldots, \mathbf{\Theta}^{[L]} \} $.
 
Predict the label with a MLP $\mathcal{F}$ that is parameterized by $\mathbf{\hat{\Theta}}$

\nlnonumber
$ \mathbf{\hat{y}}_i = \mathcal{F}(\mathbf{X}_i; \hat{\mathbf{\Theta}} | \mathbf{X}_i) $, $\mathbf{\hat{y}}_i \in \mathbb{R}$

\nlnonumber
Trainable weights: $\mathbf{\Pi}, \mathbf{\Phi}$, $\mathbf{\Theta}^{[2]}, \ldots, \mathbf{\Theta}^{[L]}$.
\caption{The \methodname Algorithm.}

\label{algo1}
\end{algorithm}

\subsection{Inferring the first layer of weights in $\mathcal{F}$ from the updated feature embeddings}\label{sec:shallow-decoder}
Finally, \methodname infers the weights in the first layer of a MLP $\mathcal{F}$ from the updated feature embeddings (Figure \ref{fig:figure1}d). In the first layer of a MLP with $h$ hidden units, each input feature $j$ corresponds to a weight vector $\mathbf{\Theta}^{[1]}_j \in \mathbb{R}^h$ (Figure \ref{fig:figure1}b). The weight matrix in the first layer of the MLP, $\mathbf{\Theta}^{[1]} \in \mathbb{R}^{d \times h}$, is simply the concatenation of $d$ weight vectors, one corresponding to each input feature. For each input feature $j$, \methodname infers the weight vector $\mathbf{\hat{\Theta}}_j^{[1]} \in \mathbb{R}^h$ from the feature embedding $\mathbf{Q}_j \in \mathbb{R}^c$ by using a shallow neural network shared across input features. Input features with similar feature embeddings will produce similar weight vectors. Thus, \methodname captures the inductive bias that input features corresponding to similar nodes in the KG should have similar corresponding weight vectors in the MLP's first layer.

\xhdr{Formal notation} \methodname infers the weight vector associated with each input feature $j$ in the first layer of $\mathcal{F}$ with
\begin{equation} \label{eq:p}\mathbf{\hat{\Theta}}^{[1]}_j  = \mathcal{B}(\mathbf{Q}_j | \mathbf{X}_i ; \mathbf{\Phi}). \end{equation}
$ \mathcal{B} $ is a shallow neural network with trainable weights $ \mathbf{\Phi}$. $ \mathbf{Q}_j $ is the updated feature embedding of $j$ which is conditioned on the specific input sample $ \mathbf{X}_i$ since the input sample is used as an input in its calculation (Section \ref{sec:adapt-feat-embeds}, Equation \ref{eq:q-message-pass}). $\mathbf{\Phi}$ are the weights of $\mathcal{B}$. $\mathcal{B}$ and its weights $\mathbf{\Phi}$ are shared for each feature $j \in \{1, \ldots, d\}$. 

\xhdr{\methodname drastically reduces the number of trainable weights compared to a standard MLP}
The sharing of $\mathcal{B}$ and $\mathbf{\Phi}$ across all input features drastically reduces the number of trainable weights compared to a standard MLP. For a high-dimensional tabular dataset (\emph{i.e.} $d \gg n$), a standard MLP $\mathcal{T}$ with $h$ hidden units has a large number of trainable weights in the first layer since $\mathbf{\Theta}^{[1]} \in \mathbb{R}^{d \times h}$. A standard MLP $\mathcal{T}$ must learn  all $dh$ of these trainable weights by backpropagation. By contrast, $\mathcal{B}$ uses a shared set of trainable weights $\mathbf{\Phi}$ to infer $\mathbf{\hat{\Theta}}_j$ from $\mathbf{Q}_j$ for every $j \in \{1, \ldots, d\}$. The number of trainable weights in $\Phi$ is small compared to $dh$ since $\mathcal{B}$ need only transform every $\mathbf{Q}_j \in \mathbb{R}^{c}$ to $\mathbf{\hat{\Theta}^{[1]}} \in \mathbb{R}^h$. Thus, $|\Phi| = ch$ (assuming $\mathcal{B}$ is a single layer neural network). $c$, the dimensionality of the feature embedding, is much less than $d$ the number of input features. As a result, $|\Phi| = ch \ll dh$ and \methodname drastically reduces the number of trainable weights in the first layer of a MLP.

\subsection{The \methodname algorithm} \methodname is outlined in Algorithm \ref{algo1}.

%% file: 400_experiment.tex
\section{Experiments}
\begin{table*}[t]
    \centering
    \caption{\textbf{\methodname outperforms statistical and deep baselines when $d \gg n$.} For every dataset, the best overall model is in \textbf{bold} and the second best model is \underline{underlined}.}
    \vspace{2mm}
    \label{tab:main-vertical}
    \adjustbox{max width=\textwidth}{%
    \begin{tabular}{l|l|cccccc}
    \toprule
\multicolumn{2}{c|}{Dataset} & MNSCLC & CM & PDAC & BRCA & CRC & CH \\ \midrule
\multicolumn{2}{c|}{\# of features $d$} & 15,390 & 13,183 & 12,932 & 12,693 & 18,206 & 19,902 \\
\midrule
\multicolumn{2}{c|}{\# of samples $n$} & 295 & 286 & 321 & 476 & 562 & 924 \\
\midrule
\multicolumn{2}{c|}{$d/n$} & 52.2 & 46.1 & 40.3 & 28.2 & 22.6 & 19.7 \\
\midrule
Classic Stat ML & Ridge & 0.153\std{0.000} & 0.390\std{0.000} & 0.344\std{0.000} & \underline{0.538\std{0.000}} & 0.376\std{0.000} & 0.546\std{0.000} \\ \midrule
Dim. Reduct. & PCA & 0.156\std{0.113} & 0.070\std{0.000} & 0.232\std{0.121} & 0.452\std{0.000} & 0.193\std{0.163} & 0.237\std{0.232} \\\midrule
\multirow{2}{*}{Feat. Select. } & LASSO & 0.168\std{0.000} & \underline{0.431\std{0.000}} & 0.346\std{0.000} & 0.470\std{0.000} & \underline{0.400\std{0.000}} & 0.547\std{0.000} \\
& STG & 0.132\std{0.130} & 0.366\std{0.043} & 0.258\std{0.055} & 0.485\std{0.037} & 0.301\std{0.010} & 0.262\std{0.076} \\ \midrule
Decision Tree & XGBoost & -0.02\std{0.000} & 0.225\std{0.000} & \underline{0.363\std{0.000}} & 0.347\std{0.000} & 0.354\std{0.000} & \underline{0.728\std{0.000}} \\ \midrule
\multirow{3}{*}{Graph Reg. } & GraphNet & 0.169\std{0.030} & 0.277\std{0.099} & 0.249\std{0.018} & 0.350\std{0.069} & 0.125\std{0.061} & 0.646\std{0.051} \\
& NC LASSO& 0.210\std{0.014} & 0.339\std{0.044} & 0.327\std{0.053} & 0.458\std{0.083} & 0.220\std{0.030} & 0.415\std{0.083} \\
& Network LASSO & 0.212\std{0.046} & 0.243\std{0.058} & 0.136\std{0.027} & 0.348\std{0.033} & 0.171\std{0.040} & 0.212\std{0.091} \\
 \midrule
Param. Infer. & Diet & -0.04\std{0.205} & 0.054\std{0.149} & 0.309\std{0.096} & 0.213\std{0.036} & 0.087\std{0.112} & 0.148\std{0.008} \\ \midrule
\multirow{4}{*}{Tabular DL } & MLP & 0.128\std{0.126} & 0.322\std{0.043} & 0.289\std{0.047} & 0.240\std{0.067} & 0.355\std{0.022} & 0.044\std{0.039} \\
& NODE & 0.003\std{0.000} & 0.150\std{0.000} & 0.190\std{0.000} & 0.512\std{0.000} & 0.344\std{0.000} & 0.181\std{0.000} \\
& TabTransformer & \underline{0.265\std{0.000}} & 0.072\std{0.000} & 0.029\std{0.000} & 0.202\std{0.000} & 0.238\std{0.000} & 0.020\std{0.000} \\
& TabNet & 0.085\std{0.028} & 0.010\std{0.068} & 0.088\std{0.037} & 0.055\std{0.037} & 0.018\std{0.016} & 0.039\std{0.026} \\ \midrule
Ours & PLATO & \bf 0.272\std{0.130} & \bf 0.435\std{0.022} & \bf 0.400\std{0.021} & \bf 0.583\std{0.019} & \bf 0.401\std{0.019} & \bf 0.770\std{0.003} \\ \bottomrule
\end{tabular}
}
\vspace{-3mm}
\end{table*}

\begin{table*}[t]
    \parbox{.53\linewidth}{
    \centering
    \caption{\textbf{$\methodname$'s performance depends on updating feature embeddings with a trainable message-passing (MP) function.}}
    \vspace{2mm}
    \adjustbox{max width=0.53\textwidth}{%
    \begin{tabular}{lcc|c}
    \toprule
    \multirow{2}{*}{Weight Infer. $\mathcal{B}$ Input} & Feature & Trainable & \multirow{2}{*}{PearsonR} \\
    & Info. & MP & \\
    \midrule 
    Updated feat. embed. $\mathbf{Q}$ &  \greencheck & \greencheck &  0.583\std{0.019}  \\
    General feat. embed $\mathbf{M}$  &  \greencheck & \redmark & 0.522\std{0.030}  \\
    None  &  \redmark & \redmark & 0.240\std{0.067}   \\
    \bottomrule
    \end{tabular}
    \label{tab:sample-adaptation}
    }
    }
    \hfill
    \parbox{.46\linewidth}{
    \centering
    \caption{\textbf{\methodname's performance depends on both feature nodes in $G$ and other nodes representing broader domain information.}}
    \vspace{2mm}
    \adjustbox{max width=0.46\textwidth}{%

    \begin{tabular}{lcc|c}
    \toprule
    \multirow{2}{*}{Auxiliary KG} & Feature & Broader & \multirow{2}{*}{PearsonR} \\
    & Info. & Info. & \\
    \midrule 
    Full KG &  \greencheck & \greencheck &  0.583\std{0.019}  \\
    Feature-only KG  &  \greencheck & \redmark & 0.539\std{0.038}  \\
    No KG  &  \redmark & \redmark & 0.240\std{0.067}   \\
    \bottomrule
    \end{tabular}
    \label{tab:subgraph}
    }
    }
    \vspace{-3mm}
\end{table*}

\begin{table*}[t]
    \parbox{.49\linewidth}{
    \centering
    \caption{\textbf{\methodname is robust to missing edges in the knowledge graph.}}
    \vspace{2mm}
    \adjustbox{max width=0.4\textwidth}{%
    \begin{tabular}{c|c}
    \toprule
    Fraction of edges in KG & PearsonR \\
    \midrule
    $100\%$ & $0.583 \pm 0.019$ \\
    $90\%$ & $0.570 \pm 0.017 $ \\
    $70\%$ & $0.537 \pm 0.044$ \\
    $50\%$ & $0.412\pm 0.011$ \\
    \bottomrule
    \end{tabular}
    \label{tab:kg-missingness}
    }
    }
    \hfill
    \parbox{.49\linewidth}{
    \centering
    \caption{\textbf{\methodname's MLP layers $2,\ldots,L$ with trainable weights are useful for performance.}}
    \vspace{2mm}
    \adjustbox{max width=0.495\textwidth}{%

    \begin{tabular}{l|p{3cm}|c}
    \toprule
    Model & Description & PearsonR \\ \midrule
    \methodname & MLP with first layer weights inferred & $0.583 \pm 0.019$ \\
    \methodname-LR & Linear regression with weights inferred & $0.550 \pm 0.020$ \\
    \bottomrule
    \end{tabular}
    \label{tab:plato-vs-graphlr}
    }
    }
    \vspace{-4mm}
\end{table*}

\begin{table*}[t]
    \centering
    \caption{\textbf{\methodname's performance is competitive with baselines when $d \sim n$.} For every dataset, the best overall model is in \textbf{bold} and the second best model is \underline{underlined}.}
    \vspace{2mm}
    \label{tab:main-bigger-data-vertical}
    \adjustbox{max width=0.825\textwidth}{%
    \begin{tabular}{l|l|cccc}
    \toprule
    \multicolumn{2}{c|}{Dataset} & ME & BC & SCLC & NSCLC \\ \midrule
    \multicolumn{2}{c|}{\# of features $d$} & 19,902 & 18,261 & 18,437 & 18,308  \\
    \midrule
    \multicolumn{2}{c|}{\# of samples $n$} & 10,064 & 10,101 & 10,712 & 16,730 \\
    \midrule
    \multicolumn{2}{c|}{$d/n$} & 2.0 & 1.8 & 1.7 & 1.1 \\ \midrule
    Classic Stat ML & Ridge & 0.566\std{0.008} & 0.483\std{0.008} & 0.604\std{0.057} & 0.679\std{0.008} \\ \midrule
    Dim. Reduct. & PCA & 0.239\std{0.310} & 0.233\std{0.294} & 0.284\std{0.274} & 0.645\std{0.000} \\ \midrule
    \multirow{2}{*}{Feat. Select. } & LASSO & 0.667\std{0.000} & 0.633\std{0.000} & 0.669\std{0.000} & 0.637\std{0.000} \\
    & STG & 0.676\std{0.000} & 0.643\std{0.000} & 0.668\std{0.000} & 0.646\std{0.000} \\ \midrule
    Decision Tree & XGBoost & \bf 0.875\std{0.000} & \underline{0.826\std{0.000}} & \underline{0.878\std{0.000}} & \bf 0.843\std{0.000} \\ \midrule
    \multirow{3}{*}{Graph Reg. } & GraphNet & 0.675\std{0.047} & 0.723\std{0.026} & 0.742\std{0.039} & 0.627\std{0.042} \\
    & NC LASSO & 0.733\std{0.016} & 0.730\std{0.027} & 0.793\std{0.009} & 0.746\std{0.023} \\
    & Network LASSO & 0.401\std{0.034} & 0.451\std{0.022} & 0.417\std{0.074} & 0.465\std{0.034} \\ \midrule
    Param. Infer. & Diet & 0.105\std{0.000} & 0.037\std{0.000} & -0.050\std{0.000} & 0.002\std{0.000} \\ \midrule
    \multirow{4}{*}{Tabular DL } & MLP & 0.487\std{0.131} & 0.508\std{0.061} & 0.537\std{0.061} & 0.573\std{0.005} \\
    & NODE & 0.870\std{0.000} & 0.420\std{0.169} & 0.801\std{0.102} & 0.487\std{0.197} \\
    & TabTransformer & 0.305\std{0.028} & 0.010\std{0.000} & 0.288\std{0.203} & 0.503\std{0.187} \\
    & TabNet & 0.667\std{0.002} & 0.624\std{0.001} & 0.657\std{0.004} & 0.647\std{0.000} \\ \midrule
    Ours & PLATO & \bf 0.875\std{0.004} & \bf 0.844\std{0.003} & \bf 0.883\std{0.002} & \underline{0.839\std{0.000}} \\ \bottomrule
    \end{tabular}
}
    \vspace{-4mm}
\end{table*}

We evaluate \methodname against \numbaselines baselines on 10 tabular datasets (\numdggndatasets with $d \gg n$, \numdsimndatasets with $d \sim n$).

\xhdr{Datasets} We use \numdggndatasets tabular $d \gg n $ datasets, \numdsimndatasets tabular $d \sim n $ datasets~\cite{gao2015high, garnett2012systematic, iorio2016landscape, yang2012genomics}, and a corresponding knowledge graph from prior studies~\cite{Luck2020, Khler2016, Kuhn2015, ruiz2021identification, Szklarczyk2020, Wishart2017, wishart2017drugbank}. The knowledge graph contains \numnodes nodes, \numedges edges, and \numrelationtypes  relation types. All datasets include features which map to a subset of knowledge graph nodes. The remaining nodes serve as broader domain knowledge. Dataset statistics are in Tables \ref{tab:main-vertical} and \ref{tab:main-bigger-data-vertical}. Further details and code are in Appendix \ref{app:dataset}.

\xhdr{Baselines}
We compare \methodname to \numbaselines state-of-the art statistical and deep baselines. We consider regularization with Ridge Regression~\cite{marquardt1975ridge}, dimensionality reduction with PCA~\cite{abdi2010principal} followed by linear regression, feature selection with LASSO \cite{tibshirani1996regression}, deep feature selection with Stochastic Gates~\cite{yamada2020feature}, and gradient boosted decision trees with XGBoost~\cite{chen2016xgboost}. We consider graph regularization on an induced subgraph of only feature nodes with GraphNet~\cite{grosenick2013interpretable}, NC LASSO~\cite{li2008network}, and Network LASSO~\cite{hallac2015network} (Appendix \ref{app:graph-reg-baselines}). We also consider tabular deep learning with a standard MLP, self-attention-based methods with TabTransformer~\cite{huang2020tabtransformer} and TabNet~\cite{arik2021tabnet}, differentiable decision trees with NODE~\cite{popov2019neural}, and weight inference with Diet Networks~\cite{romero2016diet}. We also attempted FT-Transformer~\cite{gorishniy2021revisiting}, but it experienced out of memory issues on all datasets due to the large number of features.

\xhdr{Fair Comparison of \methodname with Baselines}
To ensure a fair comparison with baselines, we follow evaluation protocols in recent tabular benchmarks~\cite{grinsztajn2022tree, gorishniy2021revisiting}. We conduct a random search with \numconfigurations configurations of every model (including \methodname) on every dataset across a broad range of hyperparameters (Appendix \ref{app:hyperparameters}). We split data with a 60/20/20 training, validation, test split. All results are computed across 3 data splits and 3 runs of each model in each data split. We report the mean and standard deviation of the Pearson correlation (PearsonR) between $\mathbf{y}$ and $\mathbf{\hat{y}}$ across runs and splits on the test set. Each model is run on a GeForce RTX 2080 TI GPU.

\subsection{Results}

\xhdr{\methodname outperforms statistical and deep baselines when $d \gg n$}
\methodname outperforms all baselines across all \numdggndatasets datasets with $d \gg n$ (Table \ref{tab:main-vertical}). \methodname achieves the largest improvement on the PDAC dataset, improving by \bestimprovement vs. XGBoost, the best baseline for PDAC (0.400 vs. 0.363). While \methodname achieves the strongest performance across all \numdggndatasets datasets,  the best performing baseline varies across datasets. Ridge Regression is the strongest baseline for BRCA, LASSO for CM and CRC, XGBoost for PDAC and CH, and TabTransformer for MNSCLC. The remaining baselines are not the strongest baseline for any dataset. We also find that the performance of a specific baseline depends largely on the dataset. TabTransformer, for example, is the best baseline for the MNSCLC dataset but the worst baseline for the CH dataset. The rank order of all models on all datasets is Appendix \ref{app:main-table-rank}.

\xhdr{\methodname's performance depends on updating feature embeddings with a trainable message-passing function}
\methodname infers the weights $\mathbf{\hat{\Theta}^{[1]}}$ in the first layer of a MLP $\mathcal{F}$ by using feature embeddings which contain prior information about the input features. \methodname first pretrains general feature embeddings $\mathbf{M} \in \mathbb{R}^{d \times c}$. \methodname then updates the feature embeddings to $\mathbf{Q} \in \mathbb{R}^{d \times c}$ with a trainable message-passing function. We test whether updating the feature embeddings based on the trainable message-passing function is necessary by evaluating \methodname's performance on the BRCA dataset in three configurations (Table \ref{tab:sample-adaptation}). The default configuration uses the updated feature embeddings $\mathbf{Q}$ generated by the message-passing function to infer $\mathbf{\hat{\Theta}}^{[1]}$ according to $\mathbf{\hat{\Theta}^{[1]}}_j = \mathcal{B}(\mathbf{Q}_j | \mathbf{X}_i)$. The second configuration uses the general feature embeddings $\mathbf{M}$ instead of $\mathbf{Q}$ to infer $\mathbf{\hat{\Theta}}^{[1]}$ according to $\mathbf{\hat{\Theta}^{[1]}}_j = \mathcal{B}(\mathbf{M}_j)$. The third configuration does not use feature embeddings and thus ablates to a standard MLP. Using general feature embeddings $\mathbf{M}$ improves over not using feature embeddings at all (0.522 vs. 0.240). Using feature embeddings $\mathbf{Q}$ that are generated by the trainable message-passing function further improves performance (0.583 vs. 0.522). Thus, updating the feature embeddings to $\mathbf{Q}$ based on the trainable message-passing function is key to \methodname's performance. 

\xhdr{\methodname's performance depends on both feature nodes and broader knowledge nodes in the auxiliary KG}
\methodname relies on an auxiliary KG $G$ which contains information describing input features and the broader domain. Information describing input features is represented as feature nodes while information describing the broader domain is represented as other nodes in $G$ (Methods \ref{sec:problem-setting}). To test the relative importance of the feature information in $G$ vs. the broader domain information, we measured the performance of \methodname on the BRCA dataset in two KG configurations: \methodname with the full KG (\emph{i.e.} both the feature nodes and the broader domain nodes) and \methodname with a ``feature-only KG'' (\emph{i.e.} an induced subgraph on only the feature nodes) (Table \ref{tab:subgraph}). We also compare to a ``No KG'' configuration in which \methodname does not have access to the KG. Without auxiliary information describing the input features or the broader domain, \methodname is ablated to a standard MLP.

We find that both the feature nodes and the broader knowledge nodes are important for $\methodname$'s performance. Using the ``feature-only KG'' configuration of \methodname improves performance vs the ``no KG'' configuration (0.539 vs 0.240). Using the ``full KG'' configuration further improves performance vs the ``feature-only KG'' configuration (0.583 vs 0.539). \methodname's performance thus relies on both the feature information and the broader domain information in the KG.

\xhdr{\methodname is robust to missing edges in the knowledge graph}
All knowledge graphs (KGs) are necessarily incomplete since there is additional knowledge to be discovered. To account for the incompleteness of the KG, \methodname uses low-dimensional embeddings from KG embedding approaches~\cite{wang2014knowledge,yang2014embedding,trouillon2016complex} which are designed to be robust to missing information, thus enabling predictive performance even with missing edges. We conduct an ablation study to assess \methodname's robustness to missing edges in the KG. We randomly remove edges from the KG and measure \methodname's performance on the BRCA dataset. We observe that with only 50\% of the KG's edges, \methodname still has 71\% of the performance as \methodname with 100\% of the KG's edges ($0.412$ vs. $0.583$) (Table \ref{tab:kg-missingness}). 

\xhdr{The importance of MLP layers $2, \ldots, L$, the layers with trainable weights, for \methodname}
\methodname is a MLP in which the weights in the first layer are inferred from the knowledge graph (KG) but the weights in the remaining layers $2, \ldots, L$ are trained normally. We conduct an ablation study to determine whether MLP layers $2, ..., L$  are necessary for \methodname's performance or whether the first layer of inferred weights are sufficient. Note that a single layer of inferred weights in \methodname is equivalent to a linear regression in which the weights are inferred from the KG. We thus compare \methodname to \methodname-LR, a linear regression in which the weights are inferred from the KG (Table \ref{tab:plato-vs-graphlr}). PLATO’s standard configuration outperforms \methodname-LR on the BRCA dataset ($0.583$ vs. $0.550$). Therefore, layers $2, \ldots, L$ of the MLP are important for \methodname's performance.

\xhdr{For datasets with $d \sim n$, \methodname is competitive with baselines}
Finally, we test $\methodname$'s performance for datasets with $d \sim n$. We test \numdsimndatasets datasets with $d \sim n$ ranging from $\frac{d}{n} = 1.1$ to $\frac{d}{n} = 2.0$ (Table \ref{tab:main-bigger-data-vertical}). We find that on \numdsimndatasets datasets with $d \sim n$, \methodname is competitive with the best performing baseline, XGBoost, but does not improve performance substantially. \methodname's stronger performance for datasets with $d \gg n$ than for datasets with $d \sim n$ is justified. \methodname's key idea is to include auxiliary information describing the input features. Auxiliary information is likely to help performance the most in settings with the least labeled data (\emph{i.e.} $d \gg n$). When $d \sim n$, auxiliary information is less helpful since the tabular dataset may already have enough information to train a strong predictive model. We further find that XGBoost is consistently the strongest baseline for datasets with $d \sim n$, in contrast to the varied performance of XGBoost on the datasets with $d \gg n$ (Table \ref{tab:main-vertical}).

%% file: 600_conclusion.tex
\section{Discussion}
\methodname achieves strong performance on tabular data when $d \gg n$ by using an auxiliary KG describing input features to regularize a multilayer perceptron (MLP) . Across \numdggndatasets datasets, \methodname outperforms \numbaselines state-of-the-art baselines by up to $\bestimprovement$. Ablations demonstrate the importance of \methodname's trainable message-passing function, of including nodes in the KG that don't represent input features but instead represent domain information, and of the layers in the MLP whose weights are trained directly rather than inferred. We also test \methodname's robustness to missing information in the KG. \methodname has several limitations. First, \methodname matches but does not improve the performance of baselines for high-dimensional datasets with more samples (\emph{i.e.} $d \sim n$). Second, \methodname depends on the existence of an auxiliary KG of domain information though future work may leverage existing methods to construct the KG from auxiliary unlabeled data~\cite{chong2020graph}. Overall, \methodname enables tabular deep learning when $d \gg n$ by using an auxiliary KG of domain information describing input features.

%% file: 700_appendix.tex
\newpage
\appendix
\onecolumn

\section{Evaluation protocol and hyperparameter ranges}\label{app:hyperparameters}
To ensure a fair comparison with baselines, we follow evaluation protocols outlined in tabular benchmarks~\cite{grinsztajn2022tree, gorishniy2021revisiting}. We conduct a random search with \numconfigurations configurations of every model (including \methodname) on every dataset across a broad range of hyperparameters. We base the hyperparameter ranges on the ranges used in prior tabular learning benchmarks~\cite{grinsztajn2022tree, gorishniy2021revisiting} and the ranges mentioned in the original papers of the methods. Hyperparameter ranges for \methodname are given in Table \ref{tab:plato_hyperparam}. Hyperparameter ranges for baseline methods are given in Table \ref{tab:baseline_hyperparam}.

\begin{table*}[h]
    \centering
    \begin{tabular}{l|l|l}
    \toprule
Module in \methodname & Hyperparameter & Range \\ \midrule
\multirow{3}{*}{ General } & Learning rate & LogUniform(1e-4, 5e-3) \\
& Batch size & [16, 32, 64] \\
& L2 & 0, LogUniform(1e-5, 1e-2) \\ \midrule
   \multirow{2}{*}{ KG $\mathcal{H}$ } & Embedding dimension $c$ & 200 \\ 
   & Embedding model & ComplEx \\ \midrule
   \multirow{3}{*}{ Message Passing (MP) $\mathcal{Q}$ } & \# Rounds $R$ & 2 \\ 
& $\beta$ & LogUniform(1e-4, 1e-1) \\
& Hidden dimension in $\mathcal{A}$ & UniformInt(16, 512) \\ \midrule
   \multirow{2}{*}{ Weight Inference $\mathcal{B}$ } & \# Layers & UniformInt(2, 6) \\ 
& Hidden dimension & UniformInt(16, 512) \\
\midrule
   \multirow{2}{*}{ Layers $2, \ldots, L$ in MLP $\mathcal{F}$} & \# Layers $L$ & UniformInt(2, 6) \\ 
& Hidden dimension & UniformInt(16, 512) \\

\bottomrule
\end{tabular}
    \caption{\textbf{Hyperparameter ranges used for \methodname.}}
    \label{tab:plato_hyperparam}
\end{table*}

\clearpage

\begin{longtable}[t]{ p{.2\textwidth} p{.25\textwidth} p{.25\textwidth}}
    \toprule
Model & Hyperparameter & Range \\ \midrule
LASSO & L1 & LogUniform(1E-4, 10) \\\midrule
Ridge & L2 & LogUniform(1E-4, 10) \\ \midrule
\multirow{13}{*}{ XGBoost } & n-estimators & UniformInt(1,2000) \\
& Max depth & UniformInt(3, 10) \\
& Min weight & LogUniform(1E-8,1E5) \\
& Subsample & Uniform(0.5, 1) \\
& Learning rate & LogUniform(1E-5,1) \\
& Col sample by level & Uniform(0.5, 1) \\
& Col sample by tree & Uniform(0.5, 1) \\
& Gamma & {0, LogUniform(1E-8, 1E2)} \\
& Lambda & {0, LogUniform(1E-8, 1E2)} \\
& Alpha & {0, LogUniform(1E-8, 1E2)} \\
& Booster & "gbtree" \\
& Early-stopping-rounds & 50 \\
& Iterations & 100 \\ \midrule
PCA & Number of PCA Components & UniformInt(2,1000) \\ \midrule
\multirow{6}{*}{ STG } & Hidden dimension & UniformInt(10, 500) \\
& Number of layers & UniformInt(1, 5) \\
& Activation & [Tanh, Relu, Sigmoid] \\
& Learning rate & LogUniform(1e-4, 1e-1) \\
& Sigma & Uniform(0.001, 2) \\
& Lambda & LogUniform(1e-3, 10) \\ \midrule
\multirow{6}{*}{ MLP } & Number of layers & UniformInt(1, 8) \\
& Hidden dimension & UniformInt(1, 512) \\
& Dropout & {0, Uniform([0,0.5])} \\
& Learning rate & LogUniform(1e-5, 1e-2) \\
& L2 & {0, LogUniform(1e-6, 1e-3)} \\
\midrule
\multirow{8}{*}{ TabNet } & Decision Steps & UniformInt(3, 10) \\
& Layer size & {2, 4, 8, 16, 32, 64} \\
& Relaxation factor & Uniform[1, 2] \\
& Sparsity loss weight & LogUniform[1e-6, 1e-1] \\
& Decay rate & Uniform[0.4, 0.95] \\
& Decay steps & {100, 500, 2000} \\
& Learning rate & Uniform(1e-3, 1e-2) \\
& Iterations & 100 \\ \midrule
\multirow{12}{*}{ TabTransformer } & Embedding dimension & {4, 8, 16, 32, 64, 128} \\
& Number of heads & UniformInt(1, 10) \\
& Number of attention blocks & UniformInt(1, 12) \\
& Attention dropout rate & Uniform(0, 0.5) \\
& Add norm dropout & Uniform(0, 0.5) \\
& Transformation activation & [Tanh, Relu, LeakyReLU] \\
& L2 & LogUniform(1e-6, 1e-1) \\
& Learning rate & LogUniform(1e-6, 1e-3) \\
& FF dropout & Uniform(0, 0.5) \\
& FF hidden multiplier & {1, 2, 3, 4, 5, 6, 7, 8, 9, 10} \\
& Out FF activation & [Tanh, Relu, LeakyReLU] \\
& Out FF dropout & Uniform(0, 0.5) \\ \midrule
\multirow{4}{*}{ NODE } & Learning rate & LogUniform(1e-5, 1) \\
& Number of layers & UniformInt(1, 10) \\
& Number of trees & UniformInt(2, 2048) \\
& Depth & UniformInt(1, 10) \\ \midrule
\multirow{6}{*}{ Diet Network } & Embedding choice & {$X^T$, random} \\
& Number of layers & UniformInt(1, 8) \\
& Hidden dimension & UniformInt(1, 512) \\
& Dropout & {0, Uniform([0,0.5])} \\
& Learning rate & LogUniform(1e-5, 1e-2) \\
& L2 & {0, LogUniform(1e-6, 1e-3)} \\\midrule
\multirow{4}{*}{ GraphNet } & Hidden dimension & UniformInt(1, 512) \\
& Learning rate & LogUniform(1e-5, 1e-2) \\
& $\lambda$ & {0, LogUniform(1e-5, 1e2)} \\
& L1 coefficient & {0, LogUniform(1e-5, 1e2)} \\\midrule
\multirow{4}{*}{ NC Lasso } & Hidden dimension & UniformInt(1, 512) \\
& Learning rate & LogUniform(1e-5, 1e-2) \\
& $\lambda$ & {0, LogUniform(1e-5, 1e2)} \\
& L1 coefficient & {0, LogUniform(1e-5, 1e2)} \\\midrule
\multirow{4}{*}{ Network Lasso } & Hidden dimension & UniformInt(1, 512) \\
& Learning rate & LogUniform(1e-5, 1e-2) \\
& $\lambda$ & {0, LogUniform(1e-5, 1e2)} \\
& L1 coefficient & {0, LogUniform(1e-5, 1e2)} \\
\bottomrule
    \caption{\textbf{Hyperparameter range for all baselines.}}
    \label{tab:baseline_hyperparam}
\end{longtable}

\clearpage

\section{Graph classification approaches} \label{app:graph_classify}
Graph classification models are not relevant for \methodname's setting. In graph classification models, every input sample is a graph with node attributes, and a model must make a prediction for that graph. The \methodname problem setting breaks fundamental assumptions made by graph classification models, rendering them not applicable. First, graph classification models assume that different samples correspond to different graphs~\cite{ying2021transformers, hu2019strategies, hu2020open}. However, in \methodname every sample corresponds to the exact same graph. There is a single background knowledge graph for all samples, and every sample has input features that correspond to the exact same nodes within the knowledge graph. Second, graph classification approaches typically assume that every node in an input graph has a node attribute~\cite{ying2021transformers, hu2019strategies, hu2020open}. However, in \methodname only a small subset of the nodes in the knowledge graph have measured feature values. Finally, graph classification approaches typically assume small graphs: the largest graph classification task in the Open Graph Benchmark has only 244 nodes~\cite{hu2020open}. However in \methodname, the knowledge graph contains \numnodes and the smallest dataset has \smallestfeaturenum features corresponding to nodes.

\section{\methodname's performance across node embedding methods for pre-training the feature embeddings}\label{sec:alternative-kg-embed}
We conduct an ablation study to assess how \methodname's performance depends on the node embedding method used to pre-train the feature embeddings (Methods \ref{sec:pretrain}). We test three shallow node embedding methods for knowledge graphs which are scalable and prominent: TransE~\cite{wang2014knowledge}, DistMult~\cite{yang2014embedding}, and ComplEx~\cite{trouillon2016complex}. We find that \methodname's performance is similar across TransE, DistMult, and ComplEx (Table \ref{tab:kg-embedding-ablations}). More generally, PLATO makes no assumption about what type of self-supervised node embedding method is used to pre-train the feature embeddings. The self-supervised embedding step is simply a module that pre-trains feature embeddings which are then passed to the message passing and weight inference modules of PLATO.

\begin{table*}[h]
    \centering
    \begin{tabular}{l|l}
    \toprule
KG Node Embedding Method & PearsonR (Test) on BRCA Dataset \\ \midrule
TransE & $0.582 \pm 0.025$ \\
DistMult & $0.575 \pm 0.011$ \\
ComplEx & $0.583 \pm 0.019$ \\
\bottomrule
\end{tabular}
    \caption{\textbf{\methodname's performance is consistent across knowledge graph node embedding methods.}}
    \label{tab:kg-embedding-ablations}
\end{table*}

\section{Rank ordering of methods for datasets with $d \gg n$}\label{app:main-table-rank}
In Table \ref{tab:main-vertical-rank}, we show the rank order performance of all models on all $d \gg n$ datasets. We find that \methodname exhibits consistent and strong performance while the performance of the baselines depends on the specific $d \gg n$ dataset. For example, TabTransformer is the second best performing of all models on the MNSCLC dataset but the worst performing of all models on the PDAC and CH datasets. Similarly, XGBoost is the second best performing of all models on PDAC but only the tenth best performing of all models on BRCA. The baselines with the most stable performance are LASSO and Ridge Regression which rank consistently between the second and eighth best of all models. 

\begin{table*}[h]
    \centering
    \caption{\textbf{For datasets with $d \gg n$, \methodname exhibits consistent and strong performance.} By contrast, the performance of the baselines varies with each dataset. For every dataset, the rank order of performance from Table \ref{tab:main-vertical} is shown. The best overall model is in \textbf{bold} and the second best model is \underline{underlined}.}
    \label{tab:main-vertical-rank}
    \adjustbox{max width=\textwidth}{%
    \begin{tabular}{l|l|cccccc}
    \toprule
\multicolumn{2}{c|}{Dataset} & MNSCLC & CM & PDAC & BRCA & CRC & CH \\ \midrule
\multicolumn{2}{c|}{\# of features $d$ / \# of samples $n$} & 52.2 & 46.1 & 40.3 & 28.2 & 22.6 & 19.7 \\ \midrule
Classic Stat ML & Ridge & 8 & 3 & 4 & \underline{2} & 3 & 5 \\ \midrule
Dim. Reduct. & PCA & 7 & 12 & 10 & 7 & 10 & 8 \\\midrule
\multirow{2}{*}{Feat. Select. } & LASSO & 6 & \underline{2} & 3 & 5 & \underline{2} & 4 \\
& STG & 9 & 4 & 8 & 4 & 7 & 7 \\ \midrule
Decision Tree & XGBoost & 14 & 9 & \underline{2} & 10 & 5 & \underline{2} \\ \midrule
\multirow{3}{*}{Graph Reg. } & GraphNet & 5 & 7 & 9 & 8 & 12 & 3 \\ 
& NC LASSO & 4 & 5 & 5 & 6 & 9 & 6 \\ 
& Network LASSO & 3 & 8 & 12 & 9 & 11 & 9 \\ \midrule
Param. Infer. & Diet & 13 & 13 & 6 & 12 & 13 & 11 \\ \midrule
\multirow{4}{*}{Tabular DL } & MLP & 10 & 6 & 7 & 11 & 4 & 12 \\
& NODE & 12 & 10 & 11 & 3 & 6 & 10 \\
& TabTransformer & \underline{2} & 11 & 14 & 13 & 8 & 14 \\
& TabNet & 11 & 14 & 13 & 14 & 14 & 13 \\ \midrule
Ours & PLATO & \bf 1 & \bf 1 & \bf 1 & \bf 1 & \bf 1 & \bf 1 \\ \bottomrule
\end{tabular}
}
\end{table*}

\begin{table*}[h]
    \centering
    \caption{\textbf{For datasets with $d \sim n$, \methodname is competitive with baselines.} For XGBoost is consistently the strongest baseline. For every dataset, the rank order of performance from Table \ref{tab:main-bigger-data-vertical-rank} is shown. The best overall model is in \textbf{bold} and the second best model is \underline{underlined}.}
    \label{tab:main-bigger-data-vertical-rank}
    \adjustbox{max width=0.85\textwidth}{%
    \begin{tabular}{l|l|cccc}
    \toprule
    \multicolumn{2}{c|}{Dataset} & ME & BC & SCLC & NSCLC \\ \midrule
    \multicolumn{2}{c|}{\# of features $d$ / \# of samples $n$} & 2.0 & 1.8 & 1.7 & 1.1 \\ \midrule
    Classic Stat ML & Ridge & 9 & 9 & 9 & 4 \\ \midrule
    Dim. Reduct. & PCA & 13 & 12 & 13 & 7 \\ \midrule
    \multirow{2}{*}{Feat. Select. } & LASSO & 7.5 & 6 & 6 & 8 \\
    & STG & 5 & 5 & 7 & 6 \\ \midrule
    Decision Tree & XGBoost & \bf 1.5 & \underline{2} & \underline{2} & \bf 1 \\ \midrule
    \multirow{3}{*}{Graph Reg. } & GraphNet & 6 & 4 & 5 & 9 \\ 
    & NC LASSO & 4 & 3 & 4 & 3 \\ 
    & Network LASSO & 11 & 10 & 11 & 13 \\ \midrule
    Param. Infer. & Diet & 14 & 13 & 14 & 14 \\ \midrule
    \multirow{4}{*}{Tabular DL } & MLP & 10 & 8 & 10 & 10 \\
    & NODE & 3 & 11 & 3 & 12 \\
    & TabTransformer & 12 & 14 & 12 & 11 \\
    & TabNet & 7.5 & 7 & 8 & 5 \\ \midrule
    Ours & PLATO & \bf 1.5 & \bf 1 & \bf 1 & \underline{2} \\ \bottomrule
    \end{tabular}
}
    
\end{table*}

\clearpage

\section{Graph regularization baselines} \label{app:graph-reg-baselines}
We test the state-of-the-art graph regularization baselines GraphNet~\cite{grosenick2013interpretable}, Network-Constrained LASSO~\cite{li2008network}, and Network LASSO~\cite{hallac2015network}. The graph regularization baselines can only consider a homogeneous graph with only features as nodes and a single edge type. For the graph regularization baselines, we thus induce a subgraph between feature nodes from the knowledge graph and collapse all edge types between feature nodes into a single edge type. In this context, GraphNet, Network-constrained LASSO, and Network LASSO correspond to a LASSO model with a mean-squared error loss and a graph regularization penalty. Let $\lambda$ be the graph regularization coefficient, $j$ and $k$ be two input features, let $E$ be the set of edges in the graph, let $\mathbf{\Theta} \in \mathbb{R}^d$ be the weights of the linear regression for $d$ input features, and let $D_j$ be the degree of feature node $j$. The graph regularization penalty for GraphNet is $\lambda \sum_{j, k \in E} (\mathbf{\Theta}_j - \mathbf{\Theta}_k)^2$, the penalty for Network-constrained LASSO is $\lambda \sum_{j, k \in E} (\frac{\mathbf{\Theta}_j}{\sqrt{D_j}} - \frac{\mathbf{\Theta}_k}{\sqrt{D_k}})^2$, and the penalty for Network LASSO is $\sum_{j, k \in E} |\mathbf{\Theta}_j - \mathbf{\Theta}_k|$.

\section{Number of trainable weights in \methodname vs. a multilayer perceptron}\label{app:trainable-weights}

\begin{table*}[h]
    \centering
    \caption{\textbf{\methodname drastically reduces the number of trainable weights compared to a multilayer perceptron (MLP) across all of the datasets.} The number of trainable weights in the best model from the hyperparameter sweep is shown for each dataset.}
    \label{tab:param-cnt-plato-vs-mlp}
    \adjustbox{max width=\textwidth}{%
    \begin{tabular}{c|cccccc|ccccc}
    \toprule
Model & MNSCLC & CM & PDAC & BRCA & CRC & CH & ME & BC & SCLC & NSCLC\\ \midrule
MLP & 429665 & 416961 & 820097 & 425217 & 200529 & 589761 & 586945 & 296113 & 298929 & 594209\\ \midrule
\methodname & 17154 & 42498 & 32066 & 17154 & 28386 & 61890 & 17154 & 28386 & 32066 & 17154 \\ \bottomrule
\end{tabular}
}
\end{table*}

\section{Dataset and code details}\label{app:dataset}
We use \numdggndatasets tabular $d \gg n $ datasets, \numdsimndatasets tabular $d \sim n $ datasets~\cite{gao2015high, garnett2012systematic, iorio2016landscape, yang2012genomics}, and a corresponding knowledge graph from prior studies~\cite{Luck2020, Khler2016, Kuhn2015, ruiz2021identification, Szklarczyk2020, Wishart2017, wishart2017drugbank}. The knowledge graph contains \numnodes nodes, \numedges edges, and \numrelationtypes relation types. Code, datasets, and the knowledge graph will be released with the final version of the paper.

\clearpage

%% file: neurips_2023.bbl
\begin{thebibliography}{10}

\bibitem{abdi2010principal}
Herv{\'e} Abdi and Lynne~J Williams.
\newblock {Principal Component Analysis}.
\newblock {\em Wiley Interdisciplinary Reviews: Computational Statistics},
  2(4):433--459, 2010.

\bibitem{ando2006learning}
Rie Ando and Tong Zhang.
\newblock Learning on graph with {Laplacian} regularization.
\newblock {\em Advances in Neural Information Processing Systems}, 19, 2006.

\bibitem{arik2021tabnet}
Sercan~{\"O} Arik and Tomas Pfister.
\newblock {TabNet}: Attentive interpretable tabular learning.
\newblock In {\em Proceedings of the AAAI Conference on Artificial
  Intelligence}, volume~35, pages 6679--6687, 2021.

\bibitem{balin2019concrete}
Muhammed~Fatih Bal{\i}n, Abubakar Abid, and James Zou.
\newblock Concrete autoencoders: Differentiable feature selection and
  reconstruction.
\newblock In {\em International Conference on Machine Learning}, pages
  444--453. Proceedings of Machine Learning Research, 2019.

\bibitem{bengio1990learning}
Yoshua Bengio, Samy Bengio, and Jocelyn Cloutier.
\newblock Learning a synaptic learning rule.
\newblock In {\em International Joint Conference on Neural Networks}, volume~2,
  pages 969--975. IEEE, 1991.

\bibitem{chen2016xgboost}
Tianqi Chen and Carlos Guestrin.
\newblock {XGBoost}: A scalable tree boosting system.
\newblock In {\em Proceedings of the SIGKDD Conference on Knowledge Discovery
  and Data Mining}, pages 785--794, 2016.

\bibitem{chong2020graph}
Yanwen Chong, Yun Ding, Qing Yan, and Shaoming Pan.
\newblock Graph-based semi-supervised learning: A review.
\newblock {\em Neurocomputing}, 408:216--230, 2020.

\bibitem{climente2019block}
H{\'e}ctor Climente-Gonz{\'a}lez, Chlo{\'e}-Agathe Azencott, Samuel Kaski, and
  Makoto Yamada.
\newblock Block {HSIC} lasso: model-free biomarker detection for ultra-high
  dimensional data.
\newblock {\em Bioinformatics}, 35(14):i427--i435, 2019.

\bibitem{deng2009imagenet}
Jia Deng, Wei Dong, Richard Socher, Li-Jia Li, Kai Li, and Li~Fei-Fei.
\newblock {ImageNet}: A large-scale hierarchical image database.
\newblock In {\em IEEE Conference on Computer Vision and Pattern Recognition},
  pages 248--255. IEEE, 2009.

\bibitem{denil2013predicting}
Misha Denil, Babak Shakibi, Laurent Dinh, Marc'Aurelio Ranzato, and Nando
  De~Freitas.
\newblock Predicting parameters in deep learning.
\newblock In {\em Advances in Neural Information Processing Systems},
  volume~26, pages 2148--2156, 2013.

\bibitem{devlin2018bert}
Jacob Devlin, Ming-Wei Chang, Kenton Lee, and Kristina Toutanova.
\newblock {BERT}: Pre-training of deep bidirectional transformers for language
  understanding.
\newblock In {\em Proceedings of the Conference of the North American Chapter
  of the Association for Computational Linguistics}, 2019.

\bibitem{d2021injecting}
Claudia d’Amato, Nicola~Flavio Quatraro, and Nicola Fanizzi.
\newblock Injecting background knowledge into embedding models for predictive
  tasks on knowledge graphs.
\newblock In {\em European Semantic Web Conference}, pages 441--457. Springer,
  2021.

\bibitem{freidling2021post}
Tobias Freidling, Benjamin Poignard, H{\'e}ctor Climente-Gonz{\'a}lez, and
  Makoto Yamada.
\newblock Post-selection inference with {HSIC-Lasso}.
\newblock In {\em International Conference on Machine Learning}, pages
  3439--3448. PMLR, 2021.

\bibitem{friedman2001greedy}
Jerome~H Friedman.
\newblock Greedy function approximation: a gradient boosting machine.
\newblock {\em Annals of Statistics}, pages 1189--1232, 2001.

\bibitem{gao2015high}
Hui Gao, Joshua~M Korn, St{\'e}phane Ferretti, John~E Monahan, Youzhen Wang,
  Mallika Singh, Chao Zhang, Christian Schnell, Guizhi Yang, Yun Zhang, et~al.
\newblock High-throughput screening using patient-derived tumor xenografts to
  predict clinical trial drug response.
\newblock {\em Nature Medicine}, 21(11):1318--1325, 2015.

\bibitem{garnett2012systematic}
Mathew~J Garnett, Elena~J Edelman, Sonja~J Heidorn, Chris~D Greenman, Anahita
  Dastur, King~Wai Lau, Patricia Greninger, I~Richard Thompson, Xi~Luo, Jorge
  Soares, et~al.
\newblock Systematic identification of genomic markers of drug sensitivity in
  cancer cells.
\newblock {\em Nature}, 483(7391):570--575, 2012.

\bibitem{gorishniy2021revisiting}
Yury Gorishniy, Ivan Rubachev, Valentin Khrulkov, and Artem Babenko.
\newblock Revisiting deep learning models for tabular data.
\newblock In {\em Advances in Neural Information Processing Systems},
  volume~34, pages 18932--18943, 2021.

\bibitem{grinsztajn2022tree}
Leo Grinsztajn, Edouard Oyallon, and Gael Varoquaux.
\newblock Why do tree-based models still outperform deep learning on typical
  tabular data?
\newblock In {\em Advances in Neural Information Processing Systems, Datasets
  and Benchmarks Track}, 2022.

\bibitem{grosenick2013interpretable}
Logan Grosenick, Brad Klingenberg, Kiefer Katovich, Brian Knutson, and
  Jonathan~E Taylor.
\newblock Interpretable whole-brain prediction analysis with {GraphNet}.
\newblock {\em NeuroImage}, 72:304--321, 2013.

\bibitem{guyon2004result}
Isabelle Guyon, Steve Gunn, Asa Ben-Hur, and Gideon Dror.
\newblock Result analysis of the {NIPS} 2003 feature selection challenge.
\newblock In {\em Advances in Neural Information Processing Systems},
  volume~17, pages 545--552, 2004.

\bibitem{ha2016hypernetworks}
David Ha, Andrew Dai, and Quoc~V Le.
\newblock Hypernetworks.
\newblock In {\em International Conference on Learning Representations}, 2016.

\bibitem{hallac2015network}
David Hallac, Jure Leskovec, and Stephen Boyd.
\newblock Network lasso: Clustering and optimization in large graphs.
\newblock In {\em Proceedings of the 21th ACM SIGKDD international conference
  on knowledge discovery and data mining}, pages 387--396, 2015.

\bibitem{hastie2009elements}
Trevor Hastie, Robert Tibshirani, Jerome~H Friedman, and Jerome~H Friedman.
\newblock {\em The elements of statistical learning: data mining, inference,
  and prediction}, volume~2.
\newblock Springer, 2009.

\bibitem{hazimeh2020tree}
Hussein Hazimeh, Natalia Ponomareva, Petros Mol, Zhenyu Tan, and Rahul
  Mazumder.
\newblock The tree ensemble layer: Differentiability meets conditional
  computation.
\newblock In {\em International Conference on Machine Learning}, pages
  4138--4148. PMLR, 2020.

\bibitem{hu2020open}
Weihua Hu, Matthias Fey, Marinka Zitnik, Yuxiao Dong, Hongyu Ren, Bowen Liu,
  Michele Catasta, and Jure Leskovec.
\newblock {Open Graph Benchmark}: Datasets for machine learning on graphs.
\newblock In {\em Advances in Neural Information Processing Systems},
  volume~33, pages 22118--22133, 2020.

\bibitem{hu2019strategies}
Weihua Hu, Bowen Liu, Joseph Gomes, Marinka Zitnik, Percy Liang, Vijay Pande,
  and Jure Leskovec.
\newblock Strategies for pre-training graph neural networks.
\newblock In {\em International Conference on Learning Representations}, 2020.

\bibitem{huang2020tabtransformer}
Xin Huang, Ashish Khetan, Milan Cvitkovic, and Zohar Karnin.
\newblock {TabTransformer}: Tabular data modeling using contextual embeddings.
\newblock {\em arXiv preprint arXiv:2012.06678}, 2020.

\bibitem{iorio2016landscape}
Francesco Iorio, Theo~A Knijnenburg, Daniel~J Vis, Graham~R Bignell, Michael~P
  Menden, Michael Schubert, Nanne Aben, Emanuel Gon{\c{c}}alves, Syd Barthorpe,
  Howard Lightfoot, et~al.
\newblock A landscape of pharmacogenomic interactions in cancer.
\newblock {\em Cell}, 166(3):740--754, 2016.

\bibitem{kasieczka2021lhc}
Gregor Kasieczka, Benjamin Nachman, David Shih, Oz~Amram, Anders Andreassen,
  Kees Benkendorder, Blaz Bortolato, Gustaaf Broojimans, Florencia Canelli,
  Jack Collins, et~al.
\newblock The {LHC Olympics} 2020: a community challenge for anomaly detection
  in high energy physics.
\newblock {\em Reports on Progress in Physics}, 84:124201, 2021.

\bibitem{ke2017lightgbm}
Guolin Ke, Qi~Meng, Thomas Finley, Taifeng Wang, Wei Chen, Weidong Ma, Qiwei
  Ye, and Tie-Yan Liu.
\newblock Light{GBM}: A highly efficient gradient boosting decision tree.
\newblock In {\em Advances in Neural Information Processing Systems},
  volume~30, pages 3146--3154, 2017.

\bibitem{Khler2016}
Sebastian K{\"o}hler, Nicole~A Vasilevsky, Mark Engelstad, Erin Foster, Julie
  McMurry, S{\'e}gol{\`e}ne Aym{\'e}, Gareth Baynam, Susan~M Bello, Cornelius~F
  Boerkoel, Kym~M Boycott, et~al.
\newblock The {Human Phenotype Ontology} in 2017.
\newblock {\em Nucleic Acids Research}, 45(D1):D865--D876, November 2016.

\bibitem{kontschieder2015deep}
Peter Kontschieder, Madalina Fiterau, Antonio Criminisi, and Samuel~Rota Bulo.
\newblock Deep neural decision forests.
\newblock In {\em Proceedings of the IEEE International Conference on Computer
  Vision}, pages 1467--1475, 2015.

\bibitem{kossen2021self}
Jannik Kossen, Neil Band, Clare Lyle, Aidan~N Gomez, Thomas Rainforth, and
  Yarin Gal.
\newblock Self-attention between datapoints: Going beyond individual
  input-output pairs in deep learning.
\newblock In {\em Advances in Neural Information Processing Systems},
  volume~34, pages 28742--28756, 2021.

\bibitem{Kuhn2015}
Michael Kuhn, Ivica Letunic, Lars~Juhl Jensen, and Peer Bork.
\newblock The {SIDER} database of drugs and side effects.
\newblock {\em Nucleic Acids Research}, 44(D1):D1075--D1079, October 2015.

\bibitem{li2008network}
Caiyan Li and Hongzhe Li.
\newblock Network-constrained regularization and variable selection for
  analysis of genomic data.
\newblock {\em Bioinformatics}, 24(9):1175--1182, 2008.

\bibitem{liu2017deep}
Bo~Liu, Ying Wei, Yu~Zhang, and Qiang Yang.
\newblock Deep neural networks for high dimension, low sample size data.
\newblock In {\em International Joint Conference on Artificial Intelligence
  Organization}, pages 2287--2293, 2017.

\bibitem{lou2017bdt}
Yin Lou and Mikhail Obukhov.
\newblock {BDT}: Gradient boosted decision tables for high accuracy and scoring
  efficiency.
\newblock In {\em Proceedings of the SIGKDD Conference on Knowledge Discovery
  and Data Mining}, pages 1893--1901, 2017.

\bibitem{lu2018deeppink}
Yang Lu, Yingying Fan, Jinchi Lv, and William Stafford~Noble.
\newblock {DeepPINK}: reproducible feature selection in deep neural networks.
\newblock In {\em Advances in Neural Information Processing Systems},
  volume~31, pages 8690--8700, 2018.

\bibitem{Luck2020}
Katja Luck, Dae-Kyum Kim, Luke Lambourne, Kerstin Spirohn, Bridget~E Begg,
  Wenting Bian, Ruth Brignall, Tiziana Cafarelli, Francisco~J Campos-Laborie,
  Benoit Charloteaux, et~al.
\newblock A reference map of the human binary protein interactome.
\newblock {\em Nature}, 580(7803):402--408, April 2020.

\bibitem{marquardt1975ridge}
Donald~W Marquardt and Ronald~D Snee.
\newblock Ridge regression in practice.
\newblock {\em The American Statistician}, 29(1):3--20, 1975.

\bibitem{meier2008group}
Lukas Meier, Sara Van De~Geer, and Peter B{\"u}hlmann.
\newblock The group {LASSO} for logistic regression.
\newblock {\em Journal of the Royal Statistical Society: Series B (Statistical
  Methodology)}, 70(1):53--71, 2008.

\bibitem{popov2019neural}
Sergei Popov, Stanislav Morozov, and Artem Babenko.
\newblock Neural oblivious decision ensembles for deep learning on tabular
  data.
\newblock In {\em International Conference on Learning Representations}, 2020.

\bibitem{prokhorenkova2018catboost}
Liudmila Prokhorenkova, Gleb Gusev, Aleksandr Vorobev, Anna~Veronika Dorogush,
  and Andrey Gulin.
\newblock {CatBoost}: unbiased boosting with categorical features.
\newblock In {\em Advances in Neural Information Processing Systems},
  volume~31, pages 6639--6649, 2018.

\bibitem{ramesh2022hierarchical}
Aditya Ramesh, Prafulla Dhariwal, Alex Nichol, Casey Chu, and Mark Chen.
\newblock Hierarchical text-conditional image generation with clip latents.
\newblock {\em arXiv:2204.06125}, 2022.

\bibitem{romero2016diet}
Adriana Romero, Pierre~Luc Carrier, Akram Erraqabi, Tristan Sylvain, Alex
  Auvolat, Etienne Dejoie, Marc-Andr{\'e} Legault, Marie-Pierre Dub{\'e},
  Julie~G Hussin, and Yoshua Bengio.
\newblock Diet networks: thin parameters for fat genomics.
\newblock In {\em International Conference on Learning Representations}, 2017.

\bibitem{ruiz2021identification}
Camilo Ruiz, Marinka Zitnik, and Jure Leskovec.
\newblock Identification of disease treatment mechanisms through the multiscale
  interactome.
\newblock {\em Nature Communications}, 12(1):1--15, 2021.

\bibitem{schmidhuber1992learning}
J{\"u}rgen Schmidhuber.
\newblock Learning to control fast-weight memories: An alternative to dynamic
  recurrent networks.
\newblock {\em Neural Computation}, 4(1):131--139, 1992.

\bibitem{shwartz2022tabular}
Ravid Shwartz-Ziv and Amitai Armon.
\newblock Tabular data: Deep learning is not all you need.
\newblock {\em Information Fusion}, 81:84--90, 2022.

\bibitem{smola2003kernels}
Alexander~J Smola and Risi Kondor.
\newblock Kernels and regularization on graphs.
\newblock In {\em Learning Theory and Kernel Machines}, pages 144--158.
  Springer, 2003.

\bibitem{somepalli2021saint}
Gowthami Somepalli, Micah Goldblum, Avi Schwarzschild, C~Bayan Bruss, and Tom
  Goldstein.
\newblock {SAINT}: Improved neural networks for tabular data via row attention
  and contrastive pre-training.
\newblock {\em arXiv preprint arXiv:2106.01342}, 2021.

\bibitem{song2019autoint}
Weiping Song, Chence Shi, Zhiping Xiao, Zhijian Duan, Yewen Xu, Ming Zhang, and
  Jian Tang.
\newblock Autoint: Automatic feature interaction learning via self-attentive
  neural networks.
\newblock In {\em Proceedings of the 28th ACM International Conference on
  Information and Knowledge Management}, pages 1161--1170, 2019.

\bibitem{Szklarczyk2020}
Damian Szklarczyk, Annika~L Gable, Katerina~C Nastou, David Lyon, Rebecca
  Kirsch, Sampo Pyysalo, Nadezhda~T Doncheva, Marc Legeay, Tao Fang, Peer Bork,
  Lars~J Jensen, and Christian von Mering.
\newblock The {STRING} database in 2021: customizable
  protein{\textendash}protein networks, and functional characterization of
  user-uploaded gene/measurement sets.
\newblock {\em Nucleic Acids Research}, 49(D1):D605--D612, November 2020.

\bibitem{tibshirani1996regression}
Robert Tibshirani.
\newblock Regression shrinkage and selection via the lasso.
\newblock {\em Journal of the Royal Statistical Society: Series B
  (Methodological)}, 58(1):267--288, 1996.

\bibitem{trouillon2016complex}
Th{\'e}o Trouillon, Johannes Welbl, Sebastian Riedel, {\'E}ric Gaussier, and
  Guillaume Bouchard.
\newblock Complex embeddings for simple link prediction.
\newblock In {\em International Conference on Machine Learning}, pages
  2071--2080. PMLR, 2016.

\bibitem{van2008visualizing}
Laurens Van~der Maaten and Geoffrey Hinton.
\newblock Visualizing data using {t-SNE}.
\newblock {\em Journal of Machine Learning Research}, 9(11):2579--2605, 2008.

\bibitem{van2009dimensionality}
Laurens Van Der~Maaten, Eric Postma, Jaap Van~den Herik, et~al.
\newblock Dimensionality reduction: a comparative.
\newblock {\em Journal of Machine Learning Research}, 10(66-71):13, 2009.

\bibitem{velivckovic2017graph}
Petar Veli{\v{c}}kovi{\'c}, Guillem Cucurull, Arantxa Casanova, Adriana Romero,
  Pietro Lio, and Yoshua Bengio.
\newblock Graph attention networks.
\newblock {\em International Conference on Learning Representations}, 2018.

\bibitem{wang2018glue}
Alex Wang, Amanpreet Singh, Julian Michael, Felix Hill, Omer Levy, and
  Samuel~R. Bowman.
\newblock {GLUE}: A multi-task benchmark and analysis platform for natural
  language understanding.
\newblock In {\em International Conference on Learning Representations}, 2019.

\bibitem{wang2017knowledge}
Quan Wang, Zhendong Mao, Bin Wang, and Li~Guo.
\newblock Knowledge graph embedding: A survey of approaches and applications.
\newblock {\em IEEE Transactions on Knowledge and Data Engineering},
  29(12):2724--2743, 2017.

\bibitem{wang2014knowledge}
Zhen Wang, Jianwen Zhang, Jianlin Feng, and Zheng Chen.
\newblock Knowledge graph embedding by translating on hyperplanes.
\newblock In {\em Proceedings of the AAAI Conference on Artificial
  Intelligence}, volume~28, pages 1112--1119, 2014.

\bibitem{Wishart2017}
David~S Wishart, Yannick~D Feunang, An~C Guo, Elvis~J Lo, Ana Marcu, Jason~R
  Grant, Tanvir Sajed, Daniel Johnson, Carin Li, Zinat Sayeeda, Nazanin
  Assempour, Ithayavani Iynkkaran, Yifeng Liu, Adam Maciejewski, Nicola Gale,
  Alex Wilson, Lucy Chin, Ryan Cummings, Diana Le, Allison Pon, Craig Knox, and
  Michael Wilson.
\newblock {DrugBank} 5.0: a major update to the {DrugBank} database for 2018.
\newblock {\em Nucleic Acids Research}, 46(D1):D1074--D1082, November 2017.

\bibitem{wishart2017drugbank}
David~S Wishart, Yannick~D Feunang, An~C Guo, Elvis~J Lo, Ana Marcu, Jason~R
  Grant, Tanvir Sajed, Daniel Johnson, Carin Li, Zinat Sayeeda, et~al.
\newblock {DrugBank} 5.0: a major update to the {DrugBank} database for 2018.
\newblock {\em Nucleic Acids Research}, 46(D1):D1074--D1082, 2017.

\bibitem{wortsman2022model}
Mitchell Wortsman, Gabriel Ilharco, Samir~Ya Gadre, Rebecca Roelofs, Raphael
  Gontijo-Lopes, Ari~S Morcos, Hongseok Namkoong, Ali Farhadi, Yair Carmon,
  Simon Kornblith, et~al.
\newblock Model soups: averaging weights of multiple fine-tuned models improves
  accuracy without increasing inference time.
\newblock In {\em International Conference on Machine Learning}, pages
  23965--23998. PMLR, 2022.

\bibitem{yamada2020feature}
Yutaro Yamada, Ofir Lindenbaum, Sahand Negahban, and Yuval Kluger.
\newblock Feature selection using stochastic gates.
\newblock In {\em International Conference on Machine Learning}, pages
  10648--10659. PMLR, 2020.

\bibitem{yang2014embedding}
Bishan Yang, Wen-tau Yih, Xiaodong He, Jianfeng Gao, and Li~Deng.
\newblock Embedding entities and relations for learning and inference in
  knowledge bases.
\newblock In {\em International Conference on Learning Representations}, 2015.

\bibitem{yang2012genomics}
Wanjuan Yang, Jorge Soares, Patricia Greninger, Elena~J Edelman, Howard
  Lightfoot, Simon Forbes, Nidhi Bindal, Dave Beare, James~A Smith, I~Richard
  Thompson, et~al.
\newblock {Genomics of Drug Sensitivity in Cancer (GDSC)}: a resource for
  therapeutic biomarker discovery in cancer cells.
\newblock {\em Nucleic Acids Research}, 41(D1):D955--D961, 2012.

\bibitem{yang2018deep}
Yongxin Yang, Irene~Garcia Morillo, and Timothy~M Hospedales.
\newblock Deep neural decision trees.
\newblock In {\em International Conference on Machine Learning, Workshop on
  Human Interpretability in Machine Learning (WHI)}, 2018.

\bibitem{ying2021transformers}
Chengxuan Ying, Tianle Cai, Shengjie Luo, Shuxin Zheng, Guolin Ke, Di~He,
  Yanming Shen, and Tie-Yan Liu.
\newblock Do transformers really perform badly for graph representation?
\newblock In {\em Advances in Neural Information Processing Systems},
  volume~34, pages 28877--28888, 2021.

\bibitem{yuan2006model}
Ming Yuan and Yi~Lin.
\newblock Model selection and estimation in regression with grouped variables.
\newblock {\em Journal of the Royal Statistical Society: Series B (Statistical
  Methodology)}, 68(1):49--67, 2006.

\end{thebibliography}
